# Computational Investigation of Low-Discrepancy Sequences in Simulation Algorithms for Bayesian Networks

Jian Cheng & Marek J. Druzdzel
Decision Systems Laboratory
School of Information Sciences and Intelligent Systems Program
University of Pittsburgh, Pittsburgh, PA 15260
{jcheng,marek}@sis.pitt.edu

## Abstract

Monte Carlo sampling has become a major vehicle for approximate inference in Bayesian networks. In this paper, we investigate a family of related simulation approaches, known collectively as quasi-Monte Carlo methods based on deterministic low-discrepancy sequences. We first outline several theoretical aspects of deterministic low-discrepancy sequences, show three examples of such sequences, and then discuss practical issues related to applying them to belief updating in Bayesian networks. We propose an algorithm for selecting direction numbers for Sobol sequence. Our experimental results show that low-discrepancy sequences (especially Sobol sequence) significantly improve the performance of simulation algorithms in Bayesian networks compared to Monte Carlo sampling.

## 1 Introduction

Since exact inference in Bayesian networks is NP-hard [Cooper, 1990], approximate inference algorithms may for very large and complex networks be the only class of algorithms that will produce any result at all. A prominent subclass of approximate algorithms is the family of schemes based on Monte Carlo sampling (also called *stochastic simulation* or *stochastic sampling* algorithms). The expected error of Monte Carlo sampling, fairly independent of the problem dimension (i.e., the number of variables involved), is of the order of $N^{-1/2}$, where $N$ is the number of samples. Random point sets generated by Monte Carlo sampling show often clusters of points and tend to take wasteful samples because of gaps in the sample space. This observation led to proposing error reduction methods by means of determinate point sets, such as low-discrepancy sequences. Low-discrepancy sequences try to utilize more uniformly distributed points. Application of low-discrepancy sequences to generation of sample points for Monte Carlo sampling leads to what is known as quasi-Monte Carlo approaches. The error bounds in quasi-Monte Carlo approaches are of the order of $(\log N)^d \cdot N^{-1}$, where $d$ is the problem dimension and $N$ is again the number of samples generated. When the number of samples is large enough, quasi-Monte Carlo methods are theoretically superior to Monte Carlo sampling. Another advantage of quasi-Monte Carlo methods is that their error bounds are deterministic.

Quasi-Monte Carlo methods have been successfully applied to computer graphics, computational physics, financial engineering, and approximate integrals (e.g., [Niederreiter, 1992a, Morokoff and Caflisch, 1995, Paskov and Traub, 1995, Papageorgiou and Traub, 1997]). They have proven their advantage in low-dimensionality problems. Even though some authors (e.g., [Bratley *et al.*, 1992, Morokoff and Caflisch, 1994]) believe that the quasi-Monte Carlo methods are not suitable for problems of high-dimensionality, tests by Paskov and Traub [1995] and Paskov [1997] have shown that quasi-Monte Carlo methods can be very effective for high-dimensional integral problems arising in computational finance. Papageorgiou and Traub [1997] have reported similarly good performance in high-dimensional integral problems arising in computational physics, demonstrating that quasi-Monte Carlo methods can be superior to Monte Carlo sampling even when the sample sizes are much smaller. These results rise the question whether quasi-Monte Carlo methods can improve sampling performance in Bayesian networks. To the best of our knowledge, application of quasi-Monte Carlo methods to Bayesian networks has not been studied before. Of particular interest here are high-dimensionality problems, i.e., Bayesian networks with a large number of variables, as these are problems that cannot be solved using exact methods.

In this paper, we investigate the advantages of applying low-discrepancy sequences to existing sampling algorithms in Bayesian networks. We first outline several theoretical aspects of deterministic quasi-Monte Carlo sequences, show three examples of low-discrepancy sequences, and then discuss practical issues related to applying them to belief updating in Bayesian networks. We propose an algorithm for selecting direction numbers for Sobol sequence. Our experimental results show that low-discrepancy sequences (especially Sobol sequence) can lead to significant performance improvements of simulation algorithms compared to existing Monte Carlo sampling algorithms.

In the following discussion, all random variables used are multiple-valued, discrete variables. Bold capital letters, such as $\mathbf{X}$, $\mathbf{A}$, denote sets of variables. Bold capital letter $\mathbf{E}$ denotes the set of evidence variables. Bold lower case letter $\mathbf{e}$, is used to denote the observations, i.e., instantiations of the set of evidence variables $\mathbf{E}$. Indexed capital letters, such as $X_i$, denote random variables. Bold lower case letter $\mathbf{a}$ denotes a particular instantiation of a set $\mathbf{A}$. $Pa(X_i)$ denotes the set of parents of node $X_i$. $\backslash$ denotes set difference. The notation for low-discrepancy sequences will be clarified as introduced in the paper.

The remainder of this paper is organized as follows. Section 2 provides a brief introduction to the concept of discrepancy and to low-discrepancy sequences. Section 3 presents construction methods for three popular low-discrepancy sequences — Halton, Sobol and Faure sequences. Section 4 discusses how these low-discrepancy sequences can be used in Bayesian networks. We also propose an algorithm for selection of direction numbers for Sobol sequence. Section 5 reports our empirical evaluation of quasi-Monte Carlo methods in Bayesian networks. Finally, Section 6 discusses the implications of our findings.

## 2 Discrepancy and Low-discrepancy Sequences

This section provides a brief introduction to the concepts of discrepancy and low-discrepancy sequences as applied in quasi-Monte Carlo methods. Our exposition is based on that of Niederreiter [1992b].

Discrepancy is a measure of nonuniformity of a sequence of points placed in a unary hypercube $[0,1]^d$. The most widely studied distance measure is the star discrepancy.

$$D_N^*(x_1,\ldots,x_N) = \sup_{0 \leq v_j < 1, j=1,\ldots,d} \left| \frac{1}{N} \sum_{i=1}^N \prod_{j=1}^d 1_{0 \leq x_i^j < v_j} - \prod_{j=1}^d v_j \right|.$$

In other words, for every subset $E$ of $[0,1]^d$ of the form $[0,v_1) \times \ldots \times [0,v_d)$, we divide the number of points $x_k$ in $E$ by $N$ and take the absolute difference between this quotient and the volume of $E$. The maximum difference is the star discrepancy $D_N^*$.

A sequence $x_1, x_2, \ldots, x_N$ of points in $[0,1]^d$ is a low-discrepancy sequence if for any $N > 1$

$$D_N^*(x_1,\ldots,x_N) \leq c(d) \cdot \frac{(\log N)^d}{N},$$

where the constant $c(d)$ depends only on the problem dimension $d$. The idea behind the low-discrepancy sequences is to let the fraction of the points within any subset $E$ of $[0,1]^d$ of the form $[0,v_1) \times \ldots \times [0,v_d)$ be as close as possible to its volume. That way, the low-discrepancy sequences will spread over $[0,1]^d$ as uniformly as possible, reducing gaps and clustering of points. Figure 1 uses two-dimensional projection of a random sequence and of a low-discrepancy sequence to demonstrate the fundamental difference between the two classes of sequences.

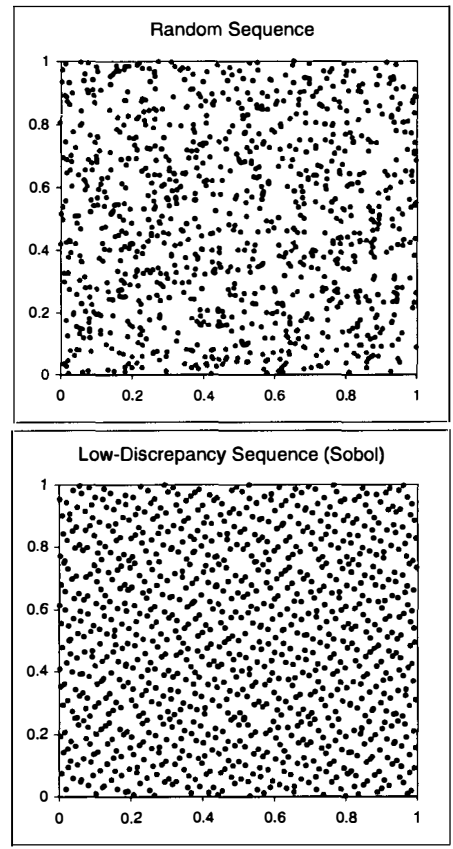

Figure 1: Two-dimensional projection of the first 1024 points in a random sequence (upper) and a low-discrepancy Sobol sequence (lower).

Suppose we want to estimate $I = \int_{[0,1]^d} f(x)dx$. Using

Monte Carlo sampling, we first generate a random sequence of independent vectors $x_1, x_2, \ldots, x_N$ from the uniform distribution on $[0,1]^d$, then use $\frac{1}{N}\sum_{i=1}^{N} f(x_i)$ as the estimator of $I$. The error bound for Monte Carlo sampling is probabilistic with order $O(N^{-1/2})$. In quasi-Monte Carlo methods, we use a low-discrepancy sequence $x_1, x_2, \ldots, x_N$ instead of a random sequence to estimate $I$. The integration accuracy for quasi-Monte Carlo methods relates to star discrepancy by the Koksma-Hlawka inequality (see [Niederreiter, 1992b])

$$\left| \int_{[0,1]^d} f(x)dx - \frac{1}{N} \sum_{i=1}^{N} f(x_i) \right|$$
$$\leq V(f) D_N^*(x_1, \ldots, x_N)$$
$$\leq V(f) c(d) \cdot \frac{(\log N)^d}{N}, \quad (1)$$

where $V(f) < \infty$ is the variation of $f$ in the sense of Hardy and Krause (see [Niederreiter, 1992b]). It is easy to see that with an increase in $N$ quasi-Monte Carlo methods may offer better convergence rates than Monte Carlo sampling. Another advantage of quasi-Monte Carlo methods is that we obtain deterministic error bounds $O((\log N)^d/N)$.

Unfortunately, the theory has it that when the problem dimension $d$ is high, it takes much larger $N$ for the quasi-Monte Carlo methods to be superior over Monte Carlo sampling. For example, in a 20-dimensional problem, $(\log N)^{20}/N$ is still greater than $1/N^{1/2}$ when $N < 10^{75}$. This observation has contributed to the belief that quasi-Monte Carlo methods are not effective in high-dimensional problems [Bratley et al., 1992, Morokoff and Caflisch, 1994]. However, several experimental tests [Paskov and Traub, 1995, Paskov, 1997, Papageorgiou and Traub, 1997] contradicted this theoretical prediction and demonstrated that quasi-Monte Carlo methods can be very effective also in high-dimensional integral problems.

The basic low-discrepancy sequences proposed in the literature are those of Halton [1960], Sobol [1967], and Faure [1982]. Niederreiter [1992b] proposed a general principle for constructing so-called $(t,d)$-sequences. The sequences of Halton, Sobol and Faure can be viewed as special cases of generalized $(t,d)$-sequences.

## 3 Construction of Low-Discrepancy Sequences

In this section, we briefly describe the construction of three low-discrepancy sequences — Halton, Sobol and Faure sequences. General principles of generating low-discrepancy sequences can be found in [Niederreiter, 1992b].

### 3.1 The Halton Sequence

Let $p_1, p_2, \ldots, p_d$ be the first $d$ prime numbers. The Halton $d$-dimensional sequence [Halton, 1960] is defined as sequence

$$x_n = (\Phi_{p_1}(n), \ldots, \Phi_{p_j}(n), \ldots, \Phi_{p_d}(n)),$$

where $\Phi_{p_j}(n)$ is the $j$th radical inverse function:

$$\Phi_{p_j}(n) = \sum_{i=0}^{l(j)} a_i(j,n) p_j^{-i-1}.$$

This sum is finite with the integer coefficients $a_i(j,n) \in [0, p_j - 1]$ ($j$ and $n$ are indexes) coming from the digit expansion of the integer $n$ in base $p_j$

$$n = \sum_{i=0}^{l(j)} a_i(j,n) p_j^i, \quad l(j) = \left\lceil \log_{p_j} n \right\rceil.$$

### 3.2 The Sobol Sequence

The Sobol sequence [1967] is generated from a set of special binary fractions of length $w$ bits, $v_i^j$, $i = 1, 2, \ldots, w$, $j = 1, 2, \ldots, d$. The numbers $v_i^j$ are called *direction numbers*.

In order to generate direction numbers for dimension $j$, we start with a primitive (irreducible) polynomial over the field $F_2$ with elements $\{0, 1\}$. Suppose the primitive polynomial in dimension $j$ is

$$p_j(x) = x^q + a_1 x^{q-1} + \ldots + a_{q-1} x + 1.$$

The direction numbers in dimension $j$ are generated using the following $q$-term recurrence relation

$$v_i^j = a_1 v_{i-1}^j \oplus a_2 v_{i-2}^j \oplus \ldots \oplus a_{q-1} v_{i-q+1}^j \oplus v_{i-q}^j \oplus (v_{i-q}^j / 2^q),$$

where $i > q$. $\oplus$ denotes the bitwise XOR operation. The initial numbers $v_1^j \cdot 2^w, v_2^j \cdot 2^w, \ldots, v_q^j \cdot 2^w$ can be arbitrary odd integers smaller than $2, 2^2, \ldots,$ and $2^q$, respectively. The Sobol sequence $x_n^j$ ($n = \sum_{i=0}^{w} b_i 2^i$, $b_i \in \{0, 1\}$) in dimension $j$ is generated by

$$x_n^j = b_1 v_1^j \oplus b_2 v_2^j \oplus \ldots \oplus b_w v_w^j.$$

We should use a different primitive polynomial to generate Sobol sequence in each dimension.

Antonov and Saleev [1979] proposed an efficient variant of Sobol sequence based on Gray code. An implementation of this variant is described in [Bratley and Fox, 1988].

### 3.3 The Faure Sequence

The Faure sequence [1982] can be generated as follows. Let $p$ be the first prime number such that $p \geq d$ and $p^m$ is the upper bound of the sample size. Let $c_{ij} = \binom{i}{j} \mod p$, $0 \le j \le i \le m$. Consider the base $p$ representation of $n$ for $n = 0, 1, 2, \ldots$,

$$n = \sum_{i=0}^{m-1} a_i(n)p^i ,$$

where $a_i(n) \in [0, p)$ are integers. The first coordinate of the point $x_n$ is then given by

$$x_n^1 = \sum_{j=0}^{m-1} a_j(n)p^{-j-1}.$$

The other coordinates are given by

$$\begin{cases} \bar{a}_j(n) = \sum_{l=j}^{m-1} c_{lj}a_l(n) \mod p, \ j \in \{0, 1, \ldots, m-1\}, \\ a_j(n) = \bar{a}_j(n), \ j \in \{0, 1, \ldots, m-1\}, \\ x_n^i = \sum_{j=0}^{m-1} a_j(n)p^{-j-1}, \end{cases}$$

in order of $i = 2, \ldots, d$. An algorithm for fast generation of Faure sequences could be found in [Tezuka, 1995].

Even though there exist other theoretical low-discrepancy sequences with asymptotically good behavior (i.e., with small value of $c(d)$), such as Niederreiter sequence [Niederreiter, 1988] or Niederreiter-Xing sequence [Niederreiter and Xing, 1996], we will not discuss them here. Practical usability of these sequences requires careful testing and solving implementational issues. It is not certain that sequences with asymptotically good behavior will necessarily perform well in practical applications, where only a finite number of points near the beginning of the sequence are used.

## 4 Quasi-Monte Carlo Methods in Bayesian Networks

In this section, we describe our adaptation of quasi-Monte Carlo algorithms to belief updating in Bayesian networks. We focus on importance sampling algorithms, currently the best performing stochastic sampling algorithms (see [Cheng and Druzdzel, 2000]). We start with a brief general description of sampling algorithms and follow this by a description of importance sampling. Finally we propose an algorithm for generation of direction numbers in Sobol sequence.

### 4.1 Stochastic Sampling in Bayesian Networks

We know that the joint probability distribution over all variables of a Bayesian network model, $\Pr(\mathbf{X})$, is the product of the probability distributions over each of the nodes conditional on their parents, i.e.,

$$\Pr(\mathbf{X}) = \prod_{i=1}^{n} \Pr(\mathbf{X}_i | \text{Pa}(\mathbf{X}_i)) .$$

In order to calculate the probability of evidence $\Pr(\mathbf{E} = \mathbf{e})$, we need to sum over all $\Pr(\mathbf{X} \setminus \mathbf{E}, \mathbf{E} = \mathbf{e})$,

$$\Pr(\mathbf{E} = \mathbf{e}) = \sum_{\mathbf{X} \setminus \mathbf{E}} \Pr(\mathbf{X} \setminus \mathbf{E}, \mathbf{E} = \mathbf{e}) . \quad (2)$$

The posterior probability $\Pr(\mathbf{a}|\mathbf{e})$ can be obtained by first computing $\Pr(\mathbf{a}, \mathbf{e})$ and $\Pr(\mathbf{e})$ separately according to equation (2), and then combining these two based on the definition of conditional probability

$$\Pr(\mathbf{a}|\mathbf{e}) = \frac{\Pr(\mathbf{a}, \mathbf{e})}{\Pr(\mathbf{e})} .$$

Stochastic sampling algorithms attempt to obtain an estimate of $\Pr(\mathbf{E} = \mathbf{e})$ in equation (2), which is analogous to approximate computation of integrals. The number of summation terms, which we will denote by $d(\mathbf{X} \setminus \mathbf{E})$, corresponds to the dimension of the problem.

In order to estimate $\Pr(\mathbf{E} = \mathbf{e})$, we can first generate a low-discrepancy sequence of $x_1, x_2, \ldots, x_N$ in $d(\mathbf{X} \setminus \mathbf{E})$ dimension unit supercube using the methods described in Section 3. Every dimension $j$ corresponds to a node in $\mathbf{X} \setminus \mathbf{E}$. Then the value $x_i^j$ can be processed as a random number generated for the corresponding node in sample $i$. Using this conversion method, low-discrepancy sequences can be easily applied to many sampling algorithms, such as probabilistic logic sampling [Henrion, 1988], likelihood weighting [Fung and Chang, 1989, Shachter and Peot, 1989], importance sampling [Shachter and Peot, 1989], or AIS-BN sampling [Cheng and Druzdzel, 2000].

### 4.2 Importance Sampling for Bayesian Networks

Sampling algorithms will in general work very well when the estimated function $\Pr(\mathbf{X} \setminus \mathbf{E}, \mathbf{E} = \mathbf{e})$ is smooth. When $\Pr(\mathbf{X} \setminus \mathbf{E}, \mathbf{E} = \mathbf{e})$ is not smooth, the performance of sampling algorithms will deteriorate (i.e., their convergence rate will be very slow). This is also true for quasi-Monte Carlo methods. Importance sampling algorithms [Shachter and Peot, 1989, Cheng and Druzdzel, 2000] address this problem by choosing an appropriate sampling distribution. The main principle of importance sampling can be summarized as an attempt to find an importance density sampling function $\Pr_{\text{isf}}(\mathbf{X} \setminus \mathbf{E})$ that will let

$$f(\mathbf{X} \setminus \mathbf{E}) = \frac{\Pr(\mathbf{X} \setminus \mathbf{E}, \mathbf{E} = \mathbf{e})}{\Pr_{\text{isf}}(\mathbf{X} \setminus \mathbf{E})} \quad (3)$$

be as smooth as possible. Another requirement for the importance sampling function $\Pr_{\text{isf}}(\mathbf{X} \setminus \mathbf{E})$ is that it should be easy to generate samples according to that function. If we generate samples $x_1, x_2, \ldots, x_N$ according to the function $\Pr_{\text{isf}}(\mathbf{X} \setminus \mathbf{E})$ independently and randomly, then $\frac{1}{N}\sum_{i=1}^{N} f(x_i)$ is an unbiased estimator of $\Pr(\mathbf{E} = \mathbf{e})$. A thorough discussions of detail of importance sampling in Bayesian networks can be found in [Cheng and Druzdzel, 2000].

### 4.3 Direction Numbers in Sobol Sequence

Suppose that we choose a primitive polynomial of degree $q$ that will generate Sobol sequence in a certain dimension. From the discussion in Section 3.2, we know that the initial numbers $v_1^j \cdot 2^w, v_2^j \cdot 2^w, \ldots, v_q^j \cdot 2^w$ in Sobol sequence can be arbitrary odd integers smaller than $2, 2^2, \ldots, 2^q$ respectively. A simple calculation shows that there are a total of $2^{q \cdot (q-1)/2}$ ways of choosing these $q$ integers. For dimension of 36 (in which case $q$ has to be 8 or higher), this number is larger than $2^{27}$. Considering all dimensions makes the total space for the initial direction numbers huge. We have found experimentally that the choice of these numbers affects the convergence rate significantly. Although Paskov and Traub [1995] and Paskov [1997] mention that they made improvements in the initial direction numbers for the Sobol sequence, they do not reveal the method that they used. This section proposes an algorithm for the choice of initial direction numbers for quasi-Monte Carlo methods in Bayesian networks.

Since the idea behind the low-discrepancy sequences is to let the points be distributed as uniformly as possible, we introduce an additional measure of uniformity of the distribution of a set of points that will be useful in choosing direction numbers. Essentially, to compute this measure of uniformity, we divide the unit square into $m^2$ equal parts. Ideally, each part should have $N/m^2$ points. We calculate the sum of the absolute differences between the actual and the ideal number of points in each part. This measure is heuristic in nature, as it looks at only two dimensions at a time. We have found empirically that the direction numbers based on this uniformity property are reasonable.

Suppose that we have obtained the initial direction numbers for the first $i$ dimensions and have derived the first $N$ points $x_l^j$, $j = 1, 2, \ldots, i$, $l = 1, 2, \ldots, N$, based on these numbers. For the dimension $i + 1$, we randomly choose the initial numbers $v_1^{i+1} \cdot 2^w, v_2^{i+1} \cdot 2^w, \ldots, v_q^{i+1} \cdot 2^w$ and then calculate $x_l^{i+1}$, $l = 1, 2, \ldots, N$. After computing the sum of the uniformity discrepancy in the unit square given by dimensions $i + 1$ and each of the $i$ dimensions based on these $N$ points, we choose the initial direction numbers that minimize the sum as our initial direction numbers in dimension $i + 1$. This is essentially a random search process. (Due to the size of the search space, it is impossible to conduct an exhaustive search.) Figure 2 contains an algorithm describing our approach.

```
for i ← 1 to nDimension do
  for j ← 1 to nRandomTimes do
    Randomly choose the initial direction numbers
        for dimension i
    S_i(N) ← Get the first N Sobol sequence in
        dimension i according to previously chosen
        initial direction numbers
    nErrorSum← 0
    for k ← 1 to i − 1 do
      nError← Calculate the uniformity discrepancy
          in the unit square given by dimensions k
          and i based on S_k(N) and S_i(N)
      nErrorSum← nErrorSum+w(k,i)·nError
    end for
    keep the best initial direction numbers and
        corresponding S_i(N) so far based on nErrorSum
  end for
end for
```

Figure 2: An algorithm for generating initial direction numbers in Sobol sequence.

We know that in Bayesian networks parent nodes affect their children directly. A good sampling heuristic is to keep parent nodes close to their children in the sampling order and to keep those dimensions that are close together more uniformly distributed. We achieve this by giving a higher weight $w(k, i)$ to those dimensions that are close to each other when computing the uniformity discrepancy. In our tests, we have chosen $N = 1,024$, $m = 32$ and $w(k, i) = 1$ when $k \geq i - 8$, otherwise $w(k, i) = 0$.

## 5 Experimental Results

We performed empirical tests comparing Monte Carlo sampling to quasi-Monte Carlo methods using five networks: COMA [Cooper, 1984], ASIA [Lauritzen and Spiegelhalter, 1988], ALARM [Beinlich et al., 1989], HAILFINDER [Abramson et al., 1996, Edwards, 1998], and a simplified version of the CPCS (Computer-based Patient Case Study) network [Pradhan et al., 1994]. The first four networks can be downloaded from http://www2.sis.pitt.edu/~genie. The CPCS network can be obtained from the Office of Technology Management, University of Pittsburgh. Each of the tested networks is multiply-connected and the last three networks are multi-layer networks with multi-valued nodes. In case of the CPCS network, we have used the largest available version for which computing the exact solution is still feasible, so that we could compute the approximation error in our experiments. The sizes of the networks (this corresponds directly to the dimension of the sampling space) ranged from 5 to 179. Each of these networks has been used in the UAI literature for the purpose of demonstration or algorithm testing. Most of them are real or realistic with both the structure and the parameters elicited from experts. We believe that our test set was quite representative for practical networks.

We focused our tests on the relationship between the number of samples and the accuracy of approximation achieved by the simulation. We measured the latter in terms of the Mean Square Error ($MSE$), i.e., square root of the sum of square differences between $\Pr'(x_{ij})$ and $\Pr(x_{ij})$, the sampled and the exact marginal probabilities of state $j$ ($j = 1, 2, \ldots, n_i$) of node $i$, such that $X_i \notin \mathbf{E}$. More precisely,

$$MSE = \sqrt{\frac{1}{\sum_{X_i \in \mathbf{X}\setminus\mathbf{E}} n_i} \sum_{X_i \in \mathbf{X}\setminus\mathbf{E}} \sum_{j=1}^{n_i} (\Pr'(x_{ij}) - \Pr(x_{ij}))^2} \;,$$

where $\mathbf{X}$ is the set of all nodes, $\mathbf{E}$ is the set of evidence nodes, and $n_i$ is the number of outcomes of node $i$. In all diagrams, the reported $MSE$ for Monte Carlo sampling is averaged over 10 runs. Since quasi-Monte Carlo methods are deterministic, we report $MSE$ of a single run.

We varied the number of samples from 250 to 256,000, staring at 250 and doubling this number at each of the subsequent 10 steps (yielding a total of 11 sampling steps in each test). In all figures included in this paper, we show plots of $\log_{10} MSE$ against $\log_2(N/250)$, where $N$ is the number of samples. We connect the points in the plots by lines in order to indicate the trend. The linear behavior observed in the log-log plots corresponds to a relationship $MSE = cN^{-\alpha}$, where $\alpha$ can be estimated by means of linear regression. For Monte Carlo sampling, the theoretical value of $\alpha$ is 0.5.

Our first tests involved belief updating without evidence. In this case, we used the probabilistic logic sampling algorithm [Henrion, 1988]. The results are shown in Figures 3 through 7. The estimated values of $\alpha$ for different networks and different sampling methods are shown in Table 1. The results of tests for low-dimensionality problems (Figures 3 and 4) show that the three quasi-Monte Carlo methods tested are significantly better than Monte Carlo sampling. The differences among the three quasi-Monte Carlo methods are small. For a given sample size, such as 8,000, the smallest improvement of $MSE$ is larger than 1,100% (one order of magnitude). The accuracy achieved by Monte Carlo sampling with 256,000 samples will be achieved by the quasi-Monte Carlo methods with only 4,000 sample points (two orders of magnitude less). With the increase of the problem dimension, the results change. The accuracy achieved by means of Faure and Halton sequences deteriorates. For the

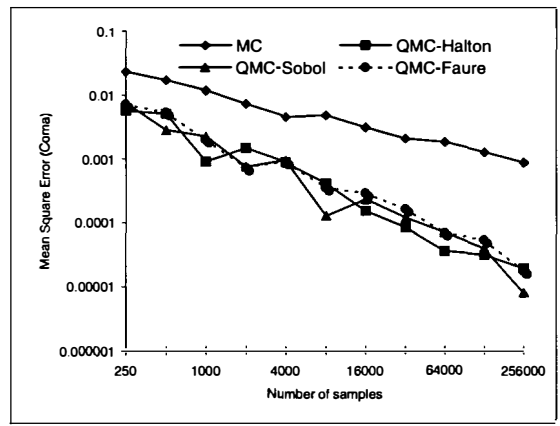

Figure 3: Mean Square Error as a function of the number of samples for the COMA network without evidence. The number of nodes in the COMA network is 5.

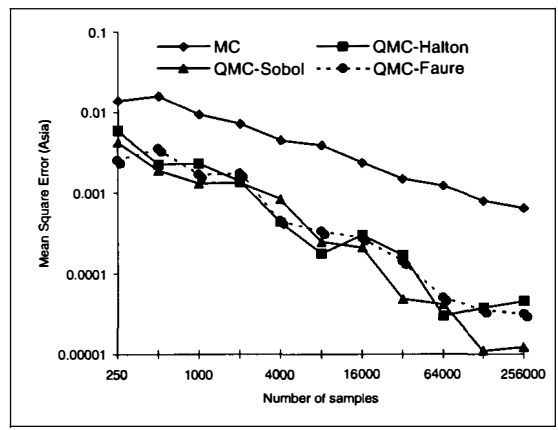

Figure 4: Mean Square Error as a function of the number of samples for the ASIA network without evidence. The number of nodes in the ASIA network is 8.

HAILFINDER and CPCS networks, the Faure sequence leads to performance that is even worse than that of Monte Carlo sampling. Although the method using Halton sequence is worse than Monte Carlo sampling when the sample size is small, its convergence rate $\alpha$ (Table 1) is better than that of Monte Carlo sampling ($\alpha = 0.5$) and when the sample size is large enough, the Halton sequence catches up. As the number of dimensions increases, the accuracy of the Sobol sequence appears to be better than that of the other two quasi-Monte Carlo methods. A remarkable result is that the method using Sobol sequence is significantly better than Monte Carlo sampling in all five tested networks. In the CPCS network, for a given sample size, such as 8,000, the improvement of $MSE$ is 372% (almost an order of magnitude). The accuracy achieved using

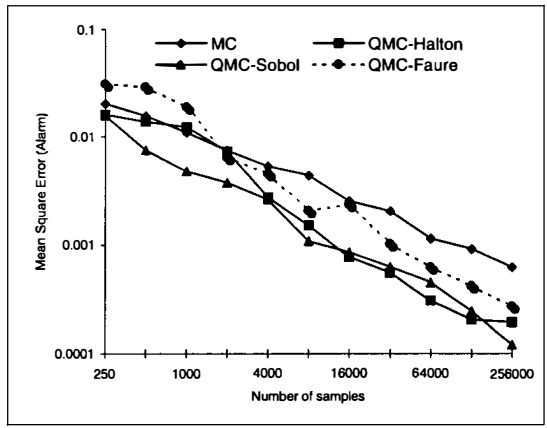

Figure 5: Mean Square Error as a function of the number of samples for the ALARM network without evidence. The number of nodes in the ALARM network is 37.

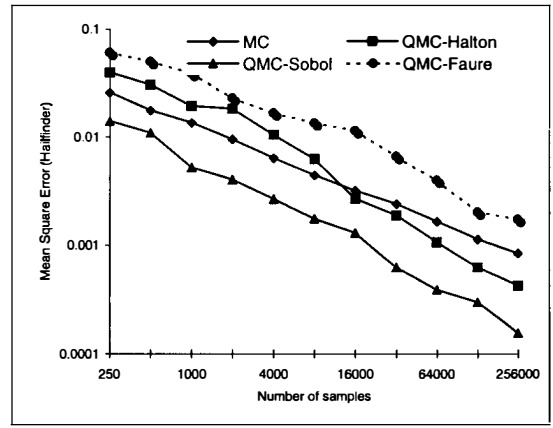

Figure 6: Mean Square Error as a function of the number of samples for the HAILFINDER network without evidence. The number of nodes in the HAILFINDER network is 56.

256,000 sample points by Monte Carlo sampling will required about 16,000 sample points in the method using Sobol sequence (still over an order of magnitude improvement). Its convergence rate $\alpha = 0.71$ is also better than that of Monte Carlo sampling ($\alpha = 0.5$), which means that higher number of samples will lead to even larger improvement. Table 1 shows that the convergence rate $\alpha$ of quasi-Monte Carlo methods was always better than Monte Carlo sampling. It seems that the method using Halton sequence leads to better convergence rates than that using Faure sequence with the increase of dimensions.

| $\alpha$ | MC | Halton | Sobol | Faure |
|---|---|---|---|---|
| COMA | 0.46 | 0.87 | 0.88 | 0.82 |
| ASIA | 0.49 | 0.76 | 0.90 | 0.74 |
| ALARM | 0.51 | 0.74 | 0.65 | 0.72 |
| HAILFINDER | 0.50 | 0.69 | 0.64 | 0.53 |
| CPCS | 0.51 | 0.74 | 0.71 | 0.57 |
| CPCS/E | 0.50 | 0.82 | 0.61 | 0.70 |

Table 1: Estimated convergence rates $\alpha$ for the six test cases and four tested sampling methods. CPCS/E stands for the CPCS network with 20 evidence nodes.

Our final test focused on evidential reasoning. As mentioned before, in evidential reasoning, when the function $\Pr(\mathbf{X}\backslash\mathbf{E}, \mathbf{E} = \mathbf{e})$ is not smooth, the performance of sampling methods will in general be poor. In order to compare Monte Carlo sampling to quasi-Monte Carlo methods, we based our tests on the adaptive importance sampling algorithm (AIS-BN) developed in our earlier work [Cheng and Druzdzel, 2000]. The AIS-BN algorithm first learns the optimal importance sampling function by adjusting dynamically equation (3). In our earlier tests involving evidential reasoning with very unlikely evidence, the AIS-BN algorithm has consistently outperformed the likelihood weighting algorithm by several orders of magnitude. As the focus of the current paper is a comparison of Monte Carlo sampling to quasi-Monte Carlo methods, we used the same importance sampling for both. We run the AIS-BN algorithm until it has found a good importance function $\Pr_{\text{isf}}(\mathbf{X}\backslash\mathbf{E})$ and then used importance sampling (Section 4.2) to compare Monte Carlo sampling to quasi-Monte Carlo methods. We used only the CPCS network in our tests, as it has observable nodes indicated as such. Our test cases for evidential reasoning were, therefore, quite realistic. Figure 8 shows a typical plot of convergence. The plot shows that the Sobol sequence leads to the best results, similarly to the results of our tests without evidence. For example, for the sample size of 8,000, the improvement of the Sobol sequence over Monte Carlo sampling is 293%. The accuracy achieved by 256,000 sample points in Monte Carlo sampling would require less than 32,000 sample points using Sobol sequence (over an order of magnitude improvement in speed over Monte Carlo sampling). Its convergence rate $\alpha = 0.61$ is better than that of Monte Carlo sampling ($\alpha = 0.5$). The behavior of Halton and Faure sequences is almost the same as without evidence. It is worth to point out that the improvement using Sobol sequence depends on the smoothness of the function $f(\mathbf{X}\backslash\mathbf{E})$ — the more smooth the function is, the higher the improvement.

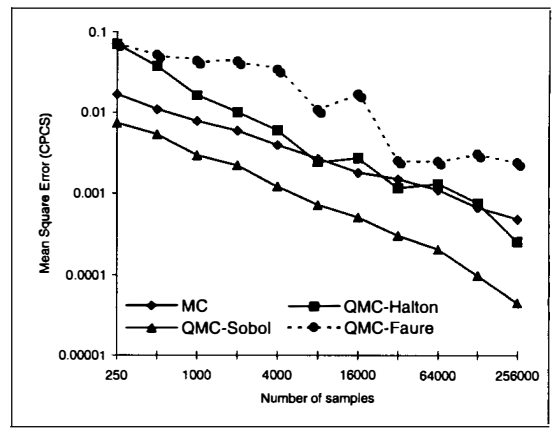

Figure 7: Mean Square Error as a function of the number of samples for a subset of the CPCS network without evidence. The number of nodes in the tested subset of the CPCS network was 179.

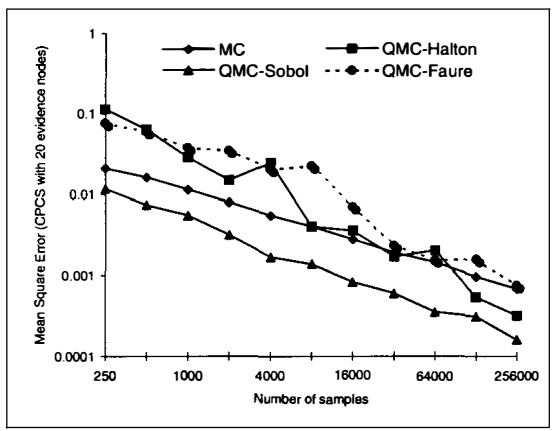

Figure 8: Mean Square Error as a function of the number of samples for the CPCS network with 20 evidence nodes chosen randomly among plausible medical observations ($\Pr(\mathbf{E} = \mathbf{e}) = 7.33 \times 10^{-20}$ in this particular case). The plot shows convergence after a smooth importance function has been identified using the AIS-BN algorithm.

Quasi-Monte Carlo methods preserve the anytime property of sampling algorithms. All plots of our experimental results indicate that the convergence curve of the Sobol sequence is quite smooth. It is fairly safe to terminate the simulation at any time and still obtain a reasonable result. This is different from Monte Carlo sampling, which is sensitive to the random seed and often shows large variance (please note that our plots of Monte Carlo sampling performance are smooth because they are averaged over 10 runs).

With an increase in problem dimension, one of the threats to accuracy is a possible significant correlation between different dimension in low-discrepancy sequences. The algorithm we used to select the direction numbers for the Sobol sequence tries to decrease this correlation. Other methods that aim at decreasing this correlation and improve the low-discrepancy sequences can be found in [Kocis and Whiten, 1997]. Although their methods are reported to reduce the error variance, we did not see significant improvement in our tests.

Currently, there exists no general rigorous theoretical justification that would explain why quasi-Monte Carlo methods are superior to Monte Carlo sampling across the variety of application studied. Several reasonable explanations have been proposed. Caflisch, Morokoff and Owen [1997] suggest that quasi-Monte Carlo methods are superior to Monte Carlo sampling if the effective dimension of the integrand is not large. Another explanation is that the error bounds $O((\log N)^d/N)$ in quasi-Monte Carlo methods are of the order of the upper bounds given by the inequality which can be a very loose inequality for a particular function. Since the inequality (1) is very conservative and calculating $V(f)$ is difficult, using inequality (1) to estimate the error is not practical. There are some papers (e.g., [Owen, 1995, Kocis and Whiten, 1997]) discussing the error estimation in quasi-Monte Carlo methods.

In terms of absolute computation time, we have observed that generation of one Sobol and one Faure point takes respectively about 57% and 29% less than generation of one random sample. As a complete sampling algorithm consists of other steps that are the same for Monte Carlo and quasi-Monte Carlo algorithms, the effective difference in computation time is smaller. We would like to caution the reader that these comparisons are implementation-dependent and an efficient algorithms for generating low-discrepancy sequences or random numbers can change these results.

## 6 Conclusion

Quasi-Monte Carlo methods can significantly improve the performance of sampling algorithms in Bayesian networks. In our tests, as the number of dimensions increased, the sampling method using Sobol sequence outperformed the methods using Halton and Faure sequences. Compared to Monte Carlo sampling, the quasi-Monte Carlo approach using Sobol sequence not only had a better start coefficient, but also had a better convergence rate. The exact improvement in performance depends on the smoothness of the sampling function. In sampling without evidence, we observed as much as a 3.5-fold improvement in the Mean Square Error. For a fixed level of the Mean Square Error, we observed more than a 15-fold decrease in sampling time. Given their consistently better performance over Monte Carlo sampling, we expect that quasi-Monte Carlo methods will be widely applied in Bayesian network inference.

We also believe that approximate inference in Bayesian networks is an excellent test bed for studying the properties of low-discrepancy sequences. There is a multitude of test data and extending the problem dimension is natural.

**Acknowledgments**

This research was supported by the National Science Foundation under Faculty Early Career Development (CAREER) Program, grant IRI-9624629, and by the Air Force Office of Scientific Research under grants F49620-97-1-0225 and F49620-00-1-0112. Malcolm Pradhan and Max Henrion of the Institute for Decision Systems Research shared with us the CPCS network with a kind permission from the developers of the Internist system at the University of Pittsburgh. Anonymous reviewers provided us with useful suggestions for improving the clarify of the paper. All experimental data have been obtained using SMILE, a Bayesian inference engine developed at the Decision Systems Laboratory and available at http://www2.sis.pitt.edu/~genie.